\documentclass[preprint]{sty/elsarticle}

\usepackage{pgfplotstable}
\usepackage{filecontents}
\usepackage{lscape}
\usepackage{amssymb}
\usepackage[first=0,last=9]{lcg}
\usepackage[table]{xcolor}
\usepackage{multicol}

\begin{document}
\begin{frontmatter}

  \title{Epigenetic opportunities for Evolutionary Computation}

  % Forgotten part of evolution computation?
  % Epigenetics: the forgotten part of evolutionary computation
  % Epigenetic, opportunity, evolutionary computation

\author[1]{Sizhe Yuen\corref{cor1}}
\ead{s.yuen@soton.ac.uk}

\author[2]{Thomas H.G. Ezard}
\ead{t.ezard@soton.ac.uk}

\author[1,3]{Adam J. Sobey}
\ead{ajs502@soton.ac.uk}

\cortext[cor1]{Corresponding author}
\address[1]{Maritime Engineering, University of Southampton,
  Southampton, SO17 1BJ, England, United Kingdom}
\address[2]{Ocean and Earth Science, National Oceanography Centre Southampton,
  European Way, University of Southampton, Southampton, SO14 3ZH, England, United Kingdom}
\address[3]{Marine and Maritime Group, Data-centric Engineering, The Alan
  Turing Institute, The British Library, London, NW1 2DB, England, United Kingdom}

\journal{Swarm and Evolutionary Computation}

\begin{abstract}
  Evolutionary Computation is a group of biologically inspired algorithms
  used to solve complex optimisation problems.
  It can be split into Evolutionary Algorithms, which
  take inspiration from genetic inheritance, and Swarm Intelligence algorithms,
  that take inspiration from cultural inheritance.
  However, recent developments have focused on
  computational or mathematical adaptions,
  leaving their biological roots behind. This has left much of the modern
  evolutionary literature relatively unexplored.
 
  To understand which evolutionary mechanisms have been considered, and which
  have been overlooked, this paper breaks down successful bio-inspired
  algorithms under a contemporary biological framework based on the Extended
  Evolutionary Synthesis, an extension of the classical,
  genetics focussed, Modern Synthesis.
  The analysis shows that Darwinism and the Modern
  Synthesis have been incorporated into Evolutionary Computation
  but that the Extended Evolutionary Synthesis has been
  broadly ignored beyond:cultural inheritance, incorporated in the sub-set
  of Swarm Intelligence algorithms, evolvability, through CMA-ES,
  and multilevel selection, through Multi-Level Selection
  Genetic Algorithm.

  The framework shows a missing gap in epigenetic inheritance for Evolutionary Computation,
  despite being a key building block in modern
  interpretations of how evolution occurs.
  Epigenetic inheritance can explain fast adaptation, without changes in an
  individual's genotype, by allowing biological organisms to self-adapt
  quickly to environmental cues, which, increases the speed of convergence
  while maintaining stability in changing environments. This leaves
  a diverse range of biologically inspired mechanisms as low hanging fruit that should
  be explored further within Evolutionary Computation.
  \end{abstract}


\end{frontmatter}

\section{Maximising the potential from biological analogies}

Evolutionary Computation has emerged as one of the most studied
branches of Artificial Intelligence. 
Hundreds of approaches have been reported over the years,
based on different bio-inspired behaviours. However, it is
difficult to determine the uniqueness of each algorithm when new approaches only
change the vocabulary and not the underlying mathematics.
The constant addition of new algorithms has led to discussions about
whether the biological analogies have gone too far
\cite{metaphor-exposed} with some arguing that chasing
new novel metaphors for algorithms risks moving
attention away from innovative ideas that make a real difference.
However, new genetic algorithms, such as cMLSGA (co-evolutionary
Multilevel Selection Genetic Algorithm) \cite{slaw-cmlsga}, have
successfully explored additional non-genetic or indirectly genetic mechanisms,
and found benefits that improve performance
on a growing number of practical problems \cite{zhenzhou-weave, slaw-real-problems}.
The key to success for Evolutionary Computation appears to be exploration of new
mechanisms without repeatedly exploring the same mechanisms under a new name.

\begin{figure}
  \centering
  \includegraphics[width=0.7\textwidth, keepaspectratio]{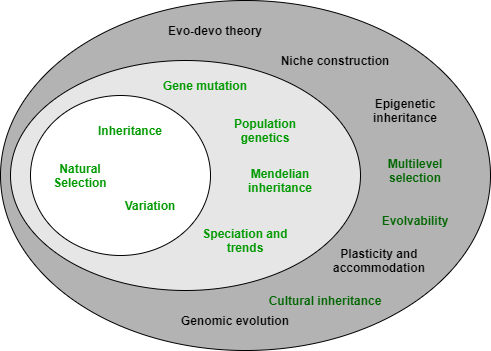}
  \caption{Key concepts of Darwinism, the Modern Synthesis, and the Extended Evolutionary Synthesis.
    Highlighted concepts are currently explored in Evolutionary Computation
    and Swarm Intelligence algorithms.}
  \label{fig:synthesis-concepts}
\end{figure}

In evolutionary theory, the Modern Synthesis \cite{modern-synthesis} was developed
throughout the first half of the $20^{th}$ century to
combine the ideas of Darwin and Wallace's evolution by natural
selection \cite{darwin-wallace, darwin:1859},
and Mendel's principles of inheritance \cite{mendel}.
Despite the contribution to evolution from these mechanisms,
there is growing evidence that suggests that
non-genetic inheritance also has a strong effect.
New research into the concepts of non-genetic inheritance suggests that
the ideas of the Modern Synthesis should be extended
\cite{pigliu-extended-synthesis, extended-synthesis-book}
to include the effects of epigenetics, cultural inheritance, parental
and environmental effects, and multilevel selection.
However, Evolutionary Computation focuses on a relatively small part
of evolutionary theory with most Evolutionary Algorithms directly
comparable to the concepts of the Modern Synthesis and Swarm Intelligence
compared directly to cultural inheritance, despite the number of algorithms
in Evolutionary Computation. Figure \ref{fig:synthesis-concepts} shows
the concepts of the Modern Synthesis and the Extended Evolutionary Synthesis,
and highlights the concepts which have been explored. While the genetic
concepts of Darwinism and the Modern Synthesis have been studied in detail,
only a few concepts from the Extended Evolutionary Synthesis have been
explored for Evolutionary Computation.
There have only been limited efforts to bring these other non-genetic inheritance
concepts to the field.

%A larger focus has been put on improving the convergence of existing
%algorithms through heuristics relevant to existing benchmark problems,
%adjusting weight vectors for concave
%problems \cite{wv-adaptive} or using predefined reference points to classify and select
%solutions in many-objective problems \cite{nsgaiii}.

There are potential benefits to investigating the outstanding evolutionary
concepts not routinely included in Evolutionary Algorithms, as demonstrated by
CMA-ES and MLSGA. In addition to those already explored there are plenty of
examples of mechanisms that have not been included such as: plasticity,
which leads to faster adaptation in the natural world \cite{lande-2009},
and maternal effects, which can increase the rate of phenotypic
evolution in changing environments while slowing it down in stable environments
to maintain optimal phenotypes \cite{ezard-maternal-effects}.
However, to help make
distinctions between different novel algorithms it is important to
understand the differences in their mechanisms , which can be
aided through a better understanding of the underlying biological concepts
and how close the new algorithms are to mimicry or inspiration.
What are the biological roots and operators
used? What are the key similarities and mathematical differences between
two bio-inspired algorithms apart from the terminology used?

To help ensure that future bio-inspired algorithms do not recycle
the same concepts, this paper categorises the current state of the art
in Evolutionary Computation under a
biological framework. The framework can show how current algorithms relate to each
other and demonstrate gaps in the evolutionary synthesis that computational algorithms
have yet to explore.

\section{Moving towards an extended evolutionary synthesis}

Despite the range of available algorithms, Evolutionary Algorithms can
be closely linked to a relatively small part of the evolutionary literature,
focused on genetic inheritance and the Modern Synthesis (Figure \ref{fig:synthesis-concepts}).
Swarm Intelligence is also linked only to the cultural inheritance
concepts of the extended synthesis. While the mechanisms from the Darwinian
concepts and the Modern Synthesis have all been included in Evolutionary Computation,
there is limited inspiration taken from the Extended Evolutionary Synthesis.

The elements of the Extended Evolutionary Synthesis expand on the existing concepts
of the Modern Synthesis to allow for a broader range of ideas and explanations
\cite{pigliu-extended-synthesis, extended-synthesis-book}.
These concepts give further explanations for how biological organisms can
adapt and change to the environment quickly compared to natural selection,
that can't be explained purely through the original concept developed by Mendel and Darwin.
In the Modern Synthesis, phenotypic variation is seen as random based
on genetic variation and inheritance. Natural selection then applies
the environmental pressure to determine the fitness of different individuals
and species. In the Extended Evolutionary Synthesis, it is argued that phenotypic
variation can be guided rather than random, allowing organisms to combat
selection pressures with fast adaptive variations to prevailing environmental conditions.
For example, evolutionary developmental biology (evo-devo) bridges the gap between
developmental biology and evolutionary biology \cite{evo-devo-history},
providing an understanding for how individual development occurs, and
how the developmental process directs phenotypic variation in individuals 
with the same genotype.
As another example, niche construction relates to ecological inheritance and
is the process in which individuals change the environmental
state to better suit their phenotypic properties \cite{niche-construction-theory}.

The different forms of non-genetic
inheritance can be divided based on their mode of transmission \cite{dachin-beyond-dna}:
\begin{itemize}
\item \textbf{Epigenetic inheritance} - In genetic inheritance, the evolution of species
  comes from changes in the DNA sequence due to natural selection. Epigenetic
  mechanisms alter the DNA expression without adapting the DNA sequence \cite{epigenetic-mice}.
  This allows
  for a rapid change in phenotypes without a change in the inherited genotype,
  leading to faster adaptation to selection pressures.
  Epigenetic marks, markers in the DNA sequence to alter the expression
  of genes, trigger changes in an individual's phenotype based on external factors such as the
  environment, and are passed on from parent to offspring.
  While epigenetic marks can cause changes to a phenotype, they
  can also remain dormant awaiting triggering by an appropriate environmental cue.
  This is a crucial difference between epigenetic and
  genetic inheritance. Traits transmitted through genetics
  are considered much harder to change or revert compared to epigenetics.
\item \textbf{Cultural inheritance} - Apart from inherited genotypes, information
  relevant to survival can also be exchanged and inherited socially. Cultures themselves
  can also ``evolve'' \cite{cultural-evolution} and can be found in a large number
  of animal species. A trait is at least partially culturally inherited if it follows
  the following four criteria \cite{dachin-inclusive-heritability}:
  \begin{enumerate}
  \item The trait is expressed as a result of learning from other individuals, and
    not inherited in another way.
  \item The trait must be passed on through multiple generations. This allows
    socially learned traits to become a part of evolution, as transmission across
    the same generation will not lead to inheritance.
  \item The phenotype of an individual must be changed as a result of social learning
    for long enough to allow new individuals to observe and learn the same trait.
  \item Changes to the individual's phenotype must be general and adaptable to other
    similar conditions.
  \end{enumerate}
\item \textbf{Ecological inheritance} - Darwin \cite{darwin-vege-mould},
  observed that individuals can alter their surrounding environment
  to improve their chances of survival. Examples of this are dams and nests
  made by animals that can sometimes be passed on to future generations. This can
  be both an in-generational, parent generation to parent generation,
  and trans-generational effect, parent generation to offspring generation, depending on whether
  the ecological changes made by one generation last for the next
  generation's benefit.
\item \textbf{Parental effects} - Parents can often have an effect on their offspring's
  phenotype during the developmental stages of the offspring's life.
  Parental effects are most often defined as the effect of parents on their offspring
  phenotype over and above the offspring's genotype. For example, some birds put
  antibodies into their eggs' yolk to defend their offspring from disease \cite{bird-antibodies}.
  Parental effects can also occur indirectly, for example in offspring seeds affected
  by the maternal light environment gaining enhanced fitness \cite{galloway-maternal-plants}.
  This directly affects the offspring's phenotype independent of their genotype;
  this is a trans-generational effect.
\item \textbf{Evolvability} - The capacity for an organism to adaptively
  evolve \cite{evolvability}. Within contemporary evolutionary biology, the
  concept takes many varied definitions, but, in algorithms, it is
  represented by hyper-parameter choices.
  The choice of mutation rate, population size, crossover method all affect
  the evolvability of the generated populations.
  Some algorithms such as CMA-ES (Covariance Matrix Adaptation Evolution Strategy)
  \cite{cma-es} and SHADE (Success-History based Parameter Adaptation
  for Differential Evolution) \cite{shade} introduce additional
  parameters and memory to dynamically evolve hyper-parameter choices
  and increase robustness to poor initial hyper-parameters.
  Evolvability also relates to genomic evolution, the evolution of genome
  architecture itself, with evolvability being the capacity
  to allow this form of evolution to occur. Computationally,
  this is related to how variables and solutions to
  problems are structured.

\end{itemize}

\section{Relating current bio-inspired algorithms to the Modern
  Extended Evolutionary Synthesis}

In Evolutionary Computation, the literature is divided between two general fields:
\begin{itemize}
\item \textbf{Evolutionary Algorithms}\cite{ec-handbook} (EC) - Algorithms with a
  foundation in genetics, these date back to Turing's learning machine
  \cite{turing:1950}, and include the branches of Evolutionary Programming (EP),
  Evolution Strategies (ES), and Genetic Algorithms (GA). These algorithms
  generally involve the genetic mechanisms of selection, recombination (crossover)
  and mutation.
\item \textbf{Swarm Intelligence} \cite{si:1999} (SI) -  Algorithms based on
  collective intelligence with patterns of communication and interaction in a
  population. Swarm Intelligence algorithms cover a wide range of biological
  inspirations, from animal behaviour algorithms such as Particle Swarm
  Optimisation \cite{pso:1995} (PSO) and Ant Colony Optimisation \cite{ant-colony}
  (ACO) as well as more esoteric inspirations such as
  political anarchy (Anarchic Society Optimisation \cite{anarchy-optimisation}).
\end{itemize}

In Evolutionary Algorithms, individuals are presented as a genotype that
evolves with genetic operators. The algorithms use genetic operators to represent
candidate solutions as genotypes and the genetic mechanisms evolve the
genotype through genetic inheritance. Conversely, individuals in
Swarm Intelligence algorithms are typically presented as phenotypes,
which are an organism's observable traits such as physical appearance
or behaviour. Phenotypes are the target for selection, for example bigger
horns, whereas genotypes represent the response to selection.
These phenotypes are often used in metaphors to hunting, foraging, and
movement behaviour, with few genetic components.
The focus of the Extended Evolutionary Synthesis is that there are
many interacting routes that influence the final phenotype,
not only genetic inheritance, as encoded in EAs, or
within-generation cultural transmission, as encoded in SI,
but rather a mix of influences both within and across
generations. This diversity of influences reduces
reliance on a single mode of inheritance
and generates phenotypes from a diversified portfolio of influences
This allows hedging against maladaptations while also providing more
rapid adaptation when genetic, indirect genetic, and phenotypic
effects align \cite{extended-synthesis-book}.

\begin{figure}
  \centering
  \includegraphics[width=0.8\textwidth, keepaspectratio]{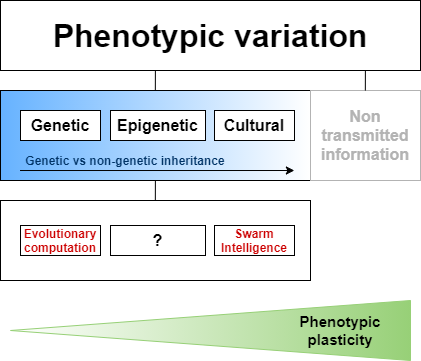}
  \caption{Sources of information transmission for phenotypic variation linked to
    inspiration for bio-inspired algorithms adapted from \cite{avatars-of-information}.}
  \label{fig:avatars-of-information}
\end{figure}

Using a biological framework for information transmission
\cite{avatars-of-information},
Figure \ref{fig:avatars-of-information} shows how the genetic and cultural sources
of information are closely linked to Evolutionary Algorithms and Swarm
Intelligence respectively. 
The mechanisms further right leverage non-genetic transmission more
than those on the left. Evolutionary Algorithms are mostly
based on the genetic source while Swarm Intelligence methods are based
on the cultural source.
Phenotypic plasticity is the ability for the same genotype to produce
different phenotypes in response to epigenetic or environmental
conditions \cite{jones-plasticity}. On the far left, genetic inheritance
has low plasticity as genotypes take many generations to mutate and evolve.
Genetic inheritance by itself cannot react to sudden changes to the environment
and adapt the genotype immediately. Cultural and non-transmitted information can lead
to higher phenotypic plasticity, higher variation, in nature as adaptation can
occur quickly within a few generations.
On the far right, non-transmitted information is information that is not inherited by
future generations; this information has the highest plasticity as
it can act immediately on environmental changes, but is unstable
and does not carry forward to future generations.
Non-transmitted information is included in some
algorithms that utilise a population of individuals acting with a certain behaviour
that does not require interaction with other individuals. For example, the Bat Algorithm
\cite{bat-algorithm} is inspired by the echolocation techniques of bats. The individual
bats interact with the environment to locate prey, but do not transmit information
with each other.

The categories of parental and ecological inheritance are omitted
for simplicity from Figure \ref{fig:avatars-of-information},
as they can be argued as a higher level of cultural
inheritance from a computational perspective.
For example, niche construction \cite{niche-construction} is the concept of changing
the environment to better suit the development of future generations. These changes
become a part of evolution when they affect natural selection, such as burrows and nests
defending family units or otherwise less fit organisms. However, niche construction
can be seen as a form of cultural
inheritance where the trait for creating ecological change in the environment is passed
down through social learning. Similarly, parental effects can be classed as part of
cultural inheritance. Parental behaviour, which affects the offspring phenotype, can
be passed on through social learning, either in-generational or trans-generational.

\begin{table}
  \caption{Biological categories of transmission for different levels of evolution.}
  \label{tbl:bio-categories}
  \begin{center}
    \begin{tabular}{| p{1.5cm} | p{3cm} | p{10cm} |}
      \hline
      \textbf{Category} & \textbf{Sub-category} & \textbf{Description} \\ \hline

      \multirow{3}{*}{Genetic}
      & Selection & In each generation, a subset of the
      population is chosen to reproduce the next generation. In biological terms,
      this relates to the survival of the fittest. Individuals with fitter genes are
      typically able to live longer and produce more offspring. In computational terms, individuals
      are selected based on their fitness against the objective function.
      \\ \cline{2-3}
      & Recombination & Also called crossover, when two individuals produce offspring,
      genetic material from both parents are used. This can be in a sexual or asexual
      way.
      This results in offspring that share some attributes with both parents.
      Crossover affects both diversity, avoiding inbreeding depression,
      and convergence of the population as it creates new
      combinations while keeping the same base genetic material from its parents.
      \\ \cline{2-3}
      & Mutation & When offspring are formed, mutation can occur.
      These mutations could be neutral, beneficial or detrimental
      to the survival of that individual. Mutation helps improve diversity by producing
      genetic material that has not been explored before by the population.
      \\ \hline

      \multirow{3}{*}{Epigenetic}
      & Mitotic & The transmission of epigenetic changes through mitotic cell divisions.
      Epigenetic marks cause variations to occur in an individual which
      are propagated through cell divisions
      when the marks are triggered as a response to environmental cues.
      Mitotic epigenetic inheritance only transfers the epigenetic
      changes across the same generation.
      \\ \cline{2-3}
      & Germline & Epigenetic changes caused by environmental factors in the parent
      are passed down to the offspring and to further generations \cite{arabadopsis-colot}.
      This affects individuals across multiple
      generations, even if the environmental factor that triggered the change only happened
      during the first generation.
      \\ \cline{2-3}
      & Experience-dependent & Epigenetic marks that influence parental behaviour causing
      the same epigenetic mark to appear in the offspring generation. The marks can persist across
      multiple generations, but the transmission can also be broken when environmental factors
      cause one generation to stop the same parental behaviour. For example, maternal
      behaviour in rodents cause the offspring to exhibit the same behaviour as their
      offspring. But if there is a break in one generation where the maternal behaviour does
      not occur, the epigenetic transmission stops \cite{champagne-rodents}.
      \\ \hline

      \multirow{3}{*}{Cultural}
      & Specialised roles & Individuals in a population have different roles to fulfil.
      For example, scout and guard bees in a colony, or mongooses that form foraging
      niches \cite{mongoose-foraging} due to competition within the population. In some
      cases, specialised roles are formed and last through multiple generations, as in
      the case of ants and bees. In others such as the mongooses, the specialised
      roles may only form in a particular generation due to external factors during
      the generation's lifespan.
      \\ \cline{2-3}
      & Social learning (individual level) & Individuals learning from other individuals
      in the population through direct information sharing, teaching, environmental
      stimulus or imitation and emulation of other individuals.
      \\ \cline{2-3}
      & Social learning (population level) & Information transfer between populations
      (cultures) where different populations may have different variations in
      behaviour as a result of social learning. 
      \\ \hline

  \end{tabular}
  \end{center}
\end{table}

\section{Categorising algorithms}

Table \ref{tbl:bio-categories-algorithms} 
breaks down each category of inheritance into the biological concepts
defined in Table \ref{tbl:bio-categories},
to show the overlap between the selected algorithms.
The algorithms were chosen if they were popular within the Evolutionary
Algorithm and Swarm Intelligence fields, defined as more than 1000 citations,
or if they showed strong performance in recent benchmarking studies
(\ref{apdx:algorithms-criteria}). There is a split between Evolutionary
Algorithms, which all use selection, recombination, and mutation operators, and
Swarm Intelligence algorithms, which have a mix between specialised roles for individuals
and different forms of social learning and communication. While most of the
algorithms fit neatly into genetic or cultural information sharing,
contrasting with evolutionary theory where multiple mechanisms
act together, there are a few
algorithms that use a spread of information mechanisms: cMLSGA uses specialised roles
and social learning, while GB-ABC and Firefly Algorithm utilise elements of selection;
GB-ABC and EBO/CMAR use recombination; HEIA uses co-evolution to spread social learning;
and Cuckoo Search uses genetically modified nests with crossover and mutation operators.
The introduction of multiple mechanisms has recently proven to balance convergence
and diversity, with cMLSGA and HEIS showing top performance on state-of-the-art
multi-objective benchmarking problems \cite{slaw-cmlsga}, and EBO/CMAR shows top
performance in a CEC'17 Single Objective Bound Constrained competition \cite{cec-17},
an area normally dominated by variants of GAs and DE algorithms.

\definecolor{ec}{rgb}{0.88,1,1}
\definecolor{si}{rgb}{1,1,0.75}

\begin{landscape}
\begin{table}[H]
  \caption{How bio-inspired algorithms fit into a biological framework for different forms
  of inheritance.}
  \label{tbl:bio-categories-algorithms}

  \begin{center}
    \begin{tabular}{| p{2cm} | c | p{1.5cm}|p{2.2cm}|p{1.5cm} | p{1.5cm}|p{1.5cm}|p{1.6cm} |
        p{1.5cm}|p{2.4cm}|p{2.4cm} |}
      \hline
      \multirow{2}{2cm}{\textbf{Algorithm category}} &
      \multirow{2}{*}{\textbf{Algorithm}} & \multicolumn{3}{c|}{\textbf{Genetic}}
      & \multicolumn{3}{c|}{\textbf{Epigenetic}} & \multicolumn{3}{c|}{\textbf{Cultural}}
      \\ \cline{3-11}
      &  & Selection & Recombination & Mutation & Mitotic & Germline & Experience-dependent &
      Specialised roles & Social learning (individual) & Social learning (population)
      \\ \hline
      \rowcolor{ec}
      & NSGA-II & \checkmark & \checkmark & \checkmark & & & & & &
      \\ \cline{2-11}
      \rowcolor{ec}
      & MOEA/D & \checkmark & \checkmark & \checkmark & & & & & & 
      \\ \cline{2-11}
      \rowcolor{ec}
      & HEIA & \checkmark & \checkmark & \checkmark & & & & \checkmark & &
      \\ \cline{2-11}
      \rowcolor{ec}
      \multirow{-4}{2cm}{Genetic Algorithm}
      & cMLSGA & \checkmark & \checkmark & \checkmark & & & & \checkmark & & \checkmark
      \\ \hline
      \rowcolor{ec}
      Evolution Strategies
      & CMA-ES & \checkmark & \checkmark & \checkmark & & & & & &
      \\ \hline
      \rowcolor{ec}
      Differential Evolution
      & SHADE & \checkmark & \checkmark & \checkmark & & & & & \checkmark &
      \\ \hline

      \rowcolor{si}
      & CLPSO & & & & & & & & \checkmark &
      \\ \cline{2-11}
      \rowcolor{si}
      & SMPSO & \checkmark & & & & & & & \checkmark &
      \\ \cline{2-11}
      \rowcolor{si}
      & ACO & & & & & & & \checkmark & \checkmark &
      \\ \cline{2-11}
      \rowcolor{si}
      & ABC & & & & & & & & \checkmark &
      \\ \cline{2-11}
      \rowcolor{si}
      & GB-ABC & \checkmark & \checkmark & & & & & \checkmark & \checkmark &
      \\ \cline{2-11}
      \rowcolor{si}
      & GWO & & & & & & & \checkmark & \checkmark &
      \\ \cline{2-11}
      \rowcolor{si}
      & EBO with CMAR & & \checkmark & & & & & \checkmark & \checkmark & \checkmark
      \\ \cline{2-11}
      \rowcolor{si}
      & Firefly Algorithm & \checkmark & & & & & & & \checkmark &
      \\ \cline{2-11}
      \rowcolor{si}
      \multirow{-8}{2cm}{Swarm Intelligence}
      & Cuckoo Search & \checkmark & * & * & & & & & * & 
      \\ \hline

    \end{tabular}
  \end{center}
\end{table}

* \textit{Cuckoo Search variants (Cuckoo-GRN and Modified Cuckoo Search) include
additional genetic and cultural mechanisms respectively.}

\begin{multicols}{2}

  NSGA-II: Non-dominated Sorting Genetic Algorithm II

  MOEA/D: Multiobjective Evolutionary Algorithm Based on Decomposition

  HEIA: Hybrid Evolutionary Immune Algorithm

  cMLSGA: Co-evolutionary Multilevel Selection Genetic Algorithm

  CMA-ES: Covariance Matrix Adaptation Evolution Strategy

  SHADE: Success-History based Parameter Adaptation for Differential Evolution

  CLPSO: Comprehensive Learning Particle Swarm Optimisation

  SMPSO: Speed-constrained Multi-objective Particle Swarm Optimisation

  ACO: Ant Colony Optimisation

  ABC: Artificial Bee Colony

  GB-ABC: Artificial Bee Colony Algorithm Based on Genetic Operators

  GWO: Grey Wolf Optimiser

  EBO with CMAR: Effective Butterfly Optimiser with Covariance Matrix Adapted Retreat Phase

\end{multicols}
\end{landscape}

While some existing studies \cite{periyasamy-ega, chrominski-ega, epiga}
investigate the use of epigenetics in Evolutionary
Computation, neither their performance nor popularity fit the criteria
to be included in this discussion.
It is proposed that this is because they do not capture the inheritance
and transfer of epigenetic information to future generations,
and so miss out the adaptability and reversibility of epigenetics.
A discussion on the key concepts of
epigenetics and why these studies miss these concepts is presented
in section \ref{sec:epigenetic-opportunities}. A justification of
Table \ref{tbl:bio-categories-algorithms} is given in the following
subsections.

\subsection{Genetic algorithms}
Genetic Algorithms (GA) are inspired by natural selection, evolving a
population of potential solutions towards an optimal solution. They are tightly
linked to a simplified concept of genetic evolution using selection,
crossover and mutation mechanics. The focus in development is almost
entirely related to different selection mechanisms, with only a small
number of modern crossover and mutation mechanisms preferred, which are no
longer bio-inspired.
In addition to these operators, many GAs also incorporate other mechanics such as
elitism \cite{elitism:1972} as a representation of ``survival of the fittest'';
which keeps the best individuals in a population,
copying them directly into the next generation to improve convergence at the cost of
diversity. This provides a considerable computational benefit and ensures that the
fittest solution can't be lost over the generations.

\subsubsection{Niching GAs}
% 4.1.1, 4.1.2, 4.1.3 all evoke ecology and eco-evolutionary interactions in some way
% could present some description to integrate the type of thinking EC is doing at the moment

The NSGA-II \cite{nsgaii} algorithm belongs to a family of niching algorithms,
where the most popular versions are NSGA-II and its refinement NSGA-III \cite{nsgaiii},
which is an extension for many-objective problems.
The difference between this approach and other Genetic Algorithms is the
use of non-dominated sorting to rank individuals
for selection but it also uses a density estimator to maintain diversity in the population.
The density estimation is not a genetic operator, as it does not involve the
genotype of candidate solutions; it is not epigenetic as it is not inherited
between individuals, and it is not cultural as it is not learned through
interaction between individuals, rather it is forced upon the population
as an external system to ensure individuals are spread out along the Pareto front.
As such it uses computational elements to improve selection but still incorporates
selection, recombination, and mutation.

\subsubsection{Decomposition based GAs}
MOEA/D \cite{moead} is a GA approach that utilises mathematical decomposition
to split a multi-objective problem into a number of sub-problems with an
assigned weight vector. The population is split to solve each sub-problem
separately, where each individual can only reproduce with individuals within
the same neighbourhood. Although the
population is divided into sub-populations and reproduction between neighbourhoods
of sub-populations is allowed, there is no learning or interaction between
individuals for social learning and this boosts exploitation of the search
but reduces the exploration. Many variants of the original MOEA/D algorithm
focus on improving different aspects of the original algorithm
such as the decomposition method \cite{mop-m2m, wfg}
or weight vector generation \cite{wv-adaptive, wv-random}. Compared to other
approaches the development of MOEA/D focuses on mathematical methods
for selection but still incorporates selection, recombination, and mutation.

\subsubsection{Co-evolutionary GAs}
Co-evolutionary GAs use cooperation or competition between two populations or
algorithms to find the best solutions. These have been gaining more popularity
in the Evolutionary Computation literature, with papers proposing that co-evolution
creates a generalist approach which reduces hyperparamter tuning \cite{slaw-real-problems}.
The top performing method is a hybrid co-evolutionary method HEIA \cite{heia}
that also uses sub-populations, but instead of dividing based
on decomposition of the problem, the sub-populations use different algorithms for
evolution, representing different selective regimes.
After each generation, the best individuals from each sub-population
are saved in an external archive and the sub-populations are cloned for the
next generation with an Immune Algorithm. With this co-evolutionary approach,
the sub-populations can naturally develop specialised roles to solve the problem.
This means the algorithm incorporates selection, recombination, mutation and
specialised roles, but with a focus on exploration of the space.

\subsubsection{Multilevel selection}
Multilevel Selection is the idea that natural selection occurs at different
levels such as the genetic level \cite{selfish-gene}, the individual organism
level \cite{nowak-kin-selection} or the species level. 
The existence of altruism, a phenotype that
contributes to group advantage at the cost of disadvantaging itself \cite{altruism},
suggests that a disadvantage at one hierarchical level may be an advantage at
another level, justifying why we see this behaviour. It does not fit into
the epigenetic and cultural concepts of non-genetic inheritance but is another
concept in the Extended Evolutionary Synthesis. MLSGA \cite{slaw-mlsga} uses the idea of
multilevel selection by evolving on both an individual level and sub-population
level (collectives). Co-evolutionary mechanics were further included in cMLSGA \cite{slaw-cmlsga}
to allow different evolutionary algorithms for the individuals of different sub-populations,
allowing collective co-evolution which increases
the generality of the algorithm. The sub-populations
are groups based on similarities between individuals, allowing specialised roles
to be formed. Further, at the sub-population level, evolution can be seen as a
form of social learning at the population level, where information is transferred
between the sub-populations. This means the algorithm incorporates selection,
recombination, mutation, specialised roles, and social learning at the population level.

\subsection{Evolution strategies}
Evolution Strategies \cite{evolution-strategies-intro}
use the mechanics of selection, recombination and mutation
to evolve the population. However, unlike traditional GAs, Evolution Strategies
are self-adapting with a distinct set of endogenous and exogenous parameters
\cite{es-introduction}. Endogenous parameters are evolved along with an individual
solution and passed down to offspring individuals, independent of others in the
population. This contrasts with GAs where all parameters such as crossover and mutation
rate are set beforehand, making them all exogenous.
Endogenous parameters control the strategy of ES algorithms
by changing the statistical properties for mutation. There are two steps of
recombination and mutation for an individual $i$ in each generation:

\begin{itemize}
\item recombination to form the endogenous strategy parameters $s(i)$
\item recombination to form the solution variables $y(i)$
\item mutation of the strategy parameters
\item mutation of the solution variables using the strategy parameters
\end{itemize}

CMA-ES \cite{cma-es} is an improvement to the traditional ES algorithm.
It uses a covariance matrix to generate its mutation distribution that is
adaptive and evolves with the search. This allows the strategy parameters to
adapt more closely during local search. The parameters are adapted based on
statistics gathered through previous generations. The self-adaptive endogenous
parameters can be seen as a form of evolvability, in the sense that
the capacity to evolve also changes \cite{evolvability}. The mechanism of passing down
and evolving this extra set of parameters changes the capacity of evolution in
the population, as the mutation rate is changed and adapted based on previous
generations. As evolvability is similar to tuning hyperparameters, the use of
a set of self-adapting parameters to control mutation automates tuning of the
mutation rate. As such it uses selection, recombination and mutation,
alongside some elements of evolvability.

\subsection{Differential evolution}
Differential evolution \cite{differential-evolution} (DE) is another branch
of evolutionary computation that is
similar to genetic algorithms. Selection, recombination and mutation
operators are all used in DE. The main difference between GAs and DE algorithms
is the representation of the individual. DE uses real value vectors
and their recombination and mutation operators revolve around the
difference between two vectors.

A popular variant to the DE algorithm, the Success-History Adaptive DE
\cite{shade} (SHADE) adapts the scaling factor and crossover rate parameters
dynamically by changing them based on a memorised history of successful
parameters in previous generations. The parameters are changed for
each individual, rather than as a whole population. This is similar to
social learning where successful parameters from other individuals
are used to evolve the parameter selection for new individuals.
This additional memory is also stored through multiple generations
for robustness, so a poor set of parameters in one generation does not
negatively impact the rest of the search. As such it uses computational
elements to improve selection, recombination and mutation alongside
some cultural elements of social learning.

\subsection{Swarm Intelligence}
In contrast to evolutionary algorithms, swarm intelligence algorithms focus
on collective behaviour and the transfer of information across the population.
Typically a single generation is used and the population traverses the search
space with different mechanics analogous to different behaviour in animals.
Some algorithms also include additional genetic components such as GB-ABC \cite{gb-abc},
an Artificial Bee Colony algorithm Based on Genetic Operators.

\subsubsection{Particle swarm optimisation}
%todo discuss smpso
Particle Swarm Optimisation (PSO) is an algorithm developed in 1995 by Eberhart and Kennedy
\cite{pso:1995} for optimising nonlinear functions. The algorithm is based on the concept of
social behaviour and sharing information. A population of particles
represents potential solutions in the search space, at each iteration, each particle
updates its velocity and position. The velocity update is based on a
combination of the best position found so far by the particle, and the best position found
so far by the entire swarm. This is an example of social
learning, as the velocity is determined by the best solution found by an individual,
but also by the best solution found so far by the entire swarm.

The Comprehensive Learning Particle Swarm Optimiser \cite{clpso} (CLPSO) extends
the idea of social learning. Rather than basing the velocity on the single
best solution found so far, CLPSO incorporates all other particles in the swarm
during the velocity update to increase diversity. The algorithm was able
to show significant improvements to performance, especially to multimodal
problems.
The PSO is based solely on social learning at the individual level
whereas CLPSO extends this to the population level.

A successful PSO algorithm for multi-objective problems is the Speed-constrained
Multi-objective PSO \cite{smpso} which uses the concepts of Pareto dominance
and a nearest neighbour density estimator to select leaders. The velocity
of particles is also limited to disallow extreme values. The algorithm is based
on social learning like other PSO algorithms, but with an added element of
selection to choose leading particles.

\subsubsection{Ant Colony Optimisation}
Ant Colony Optimisation (ACO) simulates a population of ants that moves through
the search space probabilistically based on pheromone trails left behind by
previous generations. Pheromone values decrease with each iteration so that old
trails fade and new ones form as the search space is explored.
Some pheromone trails will be reinforced if the next generations continue
to follow the same path, leading to higher pheromone values, which are associated
with better solutions. Although ACO
uses multiple iterations/generations, there is no information transfer through
genetic inheritance. The individuals in the previous generation leave behind
ecological signals with pheromone trails that affect the behaviour of new
generations, but there are no genetic operators in use. This can be seen as
ecological inheritance, which has been defined here as part of social learning.
The use of pheromone trails is core to the concept of ACO. Popular ACO algorithms
such as the Max-Min Ant System \cite{mmas} (MMAS) and the Continuous Orthogonal
Ant Colony \cite{coac} (COAC) focus on improving the mechanics of the pheromone
trails. MMAS limits the maximum and minimum pheromone values on the trails to
avoid stagnation. COAC uses orthogonal design to split up the search space into
regions for fast exploration. This was shown to improve the convergence
speed for continuous unimodal problems, finding the optimal solution in less than
half the number of function evaluations, at the cost of convergence speed for multimodal problems.
The ACO is based on social learning at the individual level and specialised roles.

\subsubsection{Artificial Bee Colony}
In an Artificial Bee Colony \cite{artificial-bees} (ABC),
the population is split into three different types of bees: employed bees, onlooker
bees and scout bees;
\begin{itemize}
\item employed bees search for better food sources in their local neighbourhood and
  share that information to onlooker bees in the region. They are able to remember
  new food sources if they are better than an existing food source in its memory.
\item Onlooker bees take the information given by employed bees and move towards
  new food sources based on the information.
\item Scout bees search for new food sources randomly without taking into account
  any information. They become employed bees when a new food source is found.
\end{itemize}
The three types of bees carry out different roles depending on their proximity to
food sources. Scout bees carry out the exploration phase of the search by moving
randomly while employed bees carry out an exploitative search by searching locally
for better food sources. Social learning can also be observed as the bees change
behaviour based on information from other bees and contextual clues from the
environment. The role of an individual bee is not static, for example onlooker
bees become employed bees after analysing the information received from other
employed bees. Similarly, employed bees become scouts when their food source is
exhausted. Some variants on the ABC algorithm include genetic operators
(GB-ABC \cite{gb-abc}) to improve global and local search for binary optimisation
problems. The ABC algorithm uses both specialised roles and social
learning between individuals with the GB-ABC variant including genetic operators.

\subsubsection{Grey Wolf Optimiser}
\begin{figure}
  \centering
  \includegraphics[width=0.7\textwidth, keepaspectratio]{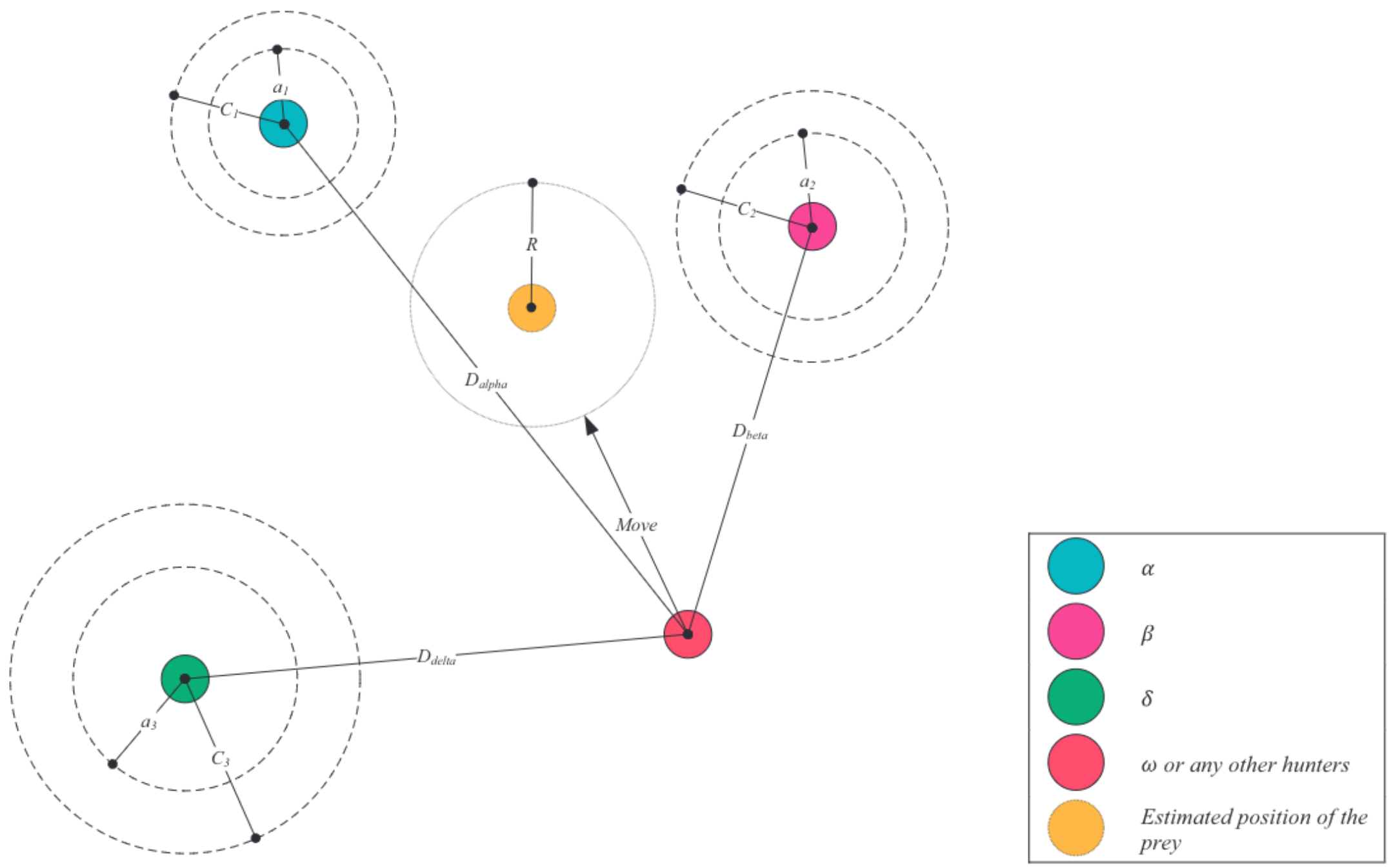}
  \caption{Position updating for the GWO algorithm.}
  \label{fig:grey-wolf}
\end{figure}

The Grey Wolf Optimizer (GWO) \cite{grey-wolf} is an algorithm inspired by grey
wolf social structure and hunting techniques. It mimics a leadership
hierarchy with four types of grey wolves, alpha, beta, delta, and omega.
During a search, the alpha, beta, and delta wolves are the three best
solutions found so far. All other individuals update their position
according to the positions of the alpha, beta, and delta wolves.
Figure \ref{fig:grey-wolf} shows how an omega wolf's position is updated
based on the positions of the alpha, beta and delta wolves.
Two coefficient vectors \textbf{\textit{A}} and \textbf{\textit{C}}
are used to fluctuate between exploration, searching for prey, and
exploitation, attacking the prey. When $|A| > 1$ the wolves diverge from
the prey instead of moving towards it. Similarly, $|C| > 1$ emphasises attacking
the prey so the wolves move faster towards it while $|C| < 1$ de-emphasises attacking.

There is a use of specialised roles among the wolf pack to form the hierarchy
and social learning occurs during a hunt, when the omega wolves follow the alpha,
beta and delta wolves for direction. Information exchange between the wolves is
used as the omega wolves move to position themselves based on the positions of
the wolves higher up in the hierarchy.
The GWO algorithm uses both specialised roles and social learning between individuals.

\subsubsection{Butterfly Optimiser}
Butterfly Optimiser \cite{butterfly-optimisation} (BO) aims to simulate the
mating behaviour of butterflies. It splits a population of butterfly individuals
into two groups, males and auxiliary butterflies in different specialised roles.
The male butterflies operate in either perching or patrolling behaviour for
exploitation and exploration of the search space respectively. Male butterflies
learn and follow auxiliary butterflies to better positions as a form of social
learning. This allows for faster exploration of the search space, as auxiliary
butterflies can continue to explore when some males decide to perch.
The attractiveness of a location affects the probability that a male butterfly
goes into perching behaviour at that location. This is an environmental factor
affecting the individual's behaviour similar to an epigenetic effect, but
because there is only a single generation, such behavioural effects are not
passed down and therefore the algorithm is not epigenetic.

An improved variant of Butterfly Algorithm, Effective Butterfly Optimiser with
Covariance Matrix Adapted Retreat Phase \cite{ebo-cmar} (EBO/CMAR) uses a
crossover operator to increase diversity and a retreat phase on a third group
of butterflies to improve convergence during local search. After each iteration
information is exchanged between each group of butterflies to replace the worst
individuals in one group with the best individuals from another. This is social
learning on a population level, as each group of butterflies uses different
behaviour and the best individuals from other groups can be learned from.
The BO uses both specialised roles and social learning
between individuals with the EBO/CMAR using genetic operators to improve the performance.

\subsubsection{Firefly Algorithm}
The Firefly Algorithm \cite{firefly-algorithm, fireflies-multimodal} (FA) is
based on the flashing patterns and behaviour of fireflies. The firefly
individuals are attracted to each other proportional to their brightness
(fitness) and the distance between two individual fireflies. Dimmer (less fit)
individuals then move towards the brighter (more fit) individuals.
A firefly's brightness is determined by the landscape of the objective function.
This means their fitness is related to an individual's phenotype rather than
their genotype. This uses elements of selection, as fireflies are more attracted
to bright individuals. Social learning is also observed, as an individual moving
towards brighter fireflies becomes brighter themselves.

\subsubsection{Cuckoo Search}
%seems to be more detailed information later on
%should we add further stuff to earlier examples or remove some stuff in later examples?
%eg, cuckoo reads like a historical diary of how the algorithm developed

Cuckoo Search \cite{cuckoo-search} is an algorithm inspired by the aggressive egg laying
behaviour of cuckoo birds, which lay eggs in the nests of other birds.
The algorithm uses the concept of Levy flights \cite{levy-flights}
to create a random walk for the population. Eggs laid in nests represent
solutions. In each iteration there is a probability for the cuckoo eggs to be
thrown out or the nest abandoned by the host birds as a selection mechanism.
In biological mechanisms, Cuckoo Search only uses selection.

However, improvements to the algorithm were made by adding either a genetic
or cultural component. In Cuckoo-GRN (Cuckoo Search with Genetically Replaced Nests)
\cite{cuckoo-grn} the convergence of the algorithm was improved by replacing
abandoned nests with crossover and mutation genetic operators. This leads to
faster convergence as new nests were created genetically using existing nests
with high fitness. Another variant, MCS (Modified Cuckoo Search) \cite{modified-cuckoo}
adds a cultural mechanism instead of a genetic mechanism. When new eggs
are generated, information is exchanged between a fraction of eggs with the
best fitness. New eggs are then generated in midpoint positions between the two
chosen eggs. The additional information sharing also improves convergence
compared to the original Cuckoo Search. These two variations demonstrate how
the addition of simple biological mechanisms helps improve performance of an algorithm.

\subsection{Summary of current algorithms in a biological framework}
\begin{figure}[H]
  \centering
  \includegraphics[width=0.7\textwidth, keepaspectratio]{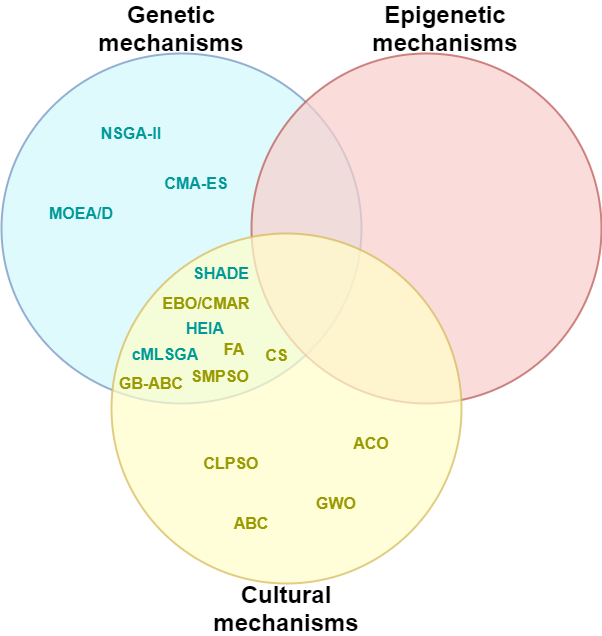}
  \caption{How bio-inspired algorithms fit into biological sources for phenotypic variation.}
  \label{fig:concept-venn-diagram}
\end{figure}

Figure \ref{fig:concept-venn-diagram} categorises the mechanisms of the main families
of Evolutionary Algorithms and Swarm Intelligence under either genetic, epigenetic
or cultural information transfer, using a single algorithm to represent each family.
%The algorithms were chosen if they were well known and popular within the Evolutionary
%Computation and Swarm Intelligence fields, defined as more than 1000 citations,
%or if they showed
%strong performance recent studies and competitions (\ref{apdx:algorithms-criteria}).
In general, most of the categories of algorithm use a single category of information
transfer, the Evolutionary Algorithms focus on different forms of genetic transfer
with mechanisms focused around changing the selection of the population to mate
and Swarm Intelligence focuses on different forms of cultural
information transfer. There are currently no successful epigenetic
mechanisms, despite it's importance in describing the process of evolution.
Many of the popular Evolutionary Algorithms
such as NSGA-II \cite{nsgaii}, MOEA/D \cite{moead}, and SHADE \cite{shade}
have diverged from their biological roots
into mathematical or statistical methods.

However, there are algorithms that show some overlap between genetic and cultural
mechanisms, which can be seen for both Evolutionary Algorithms and Swarm Intelligence
algorithms. For example, cMLSGA uses multilevel
selection by evolving both the individuals and the collectives of individuals, and
GB-ABC \cite{gb-abc} adds genetic operators to the bee colony algorithm.
The mechanisms resulted in increased diversity for cMLSGA and increased convergence for
GB-ABC.
The main gap is in epigenetic transfer with only EBO/CMAR (Effective Butterfly Optimiser
with Covariance Matrix Adapted Retreat Phase), using elements of this type of
information transfer but which still can't be considered to be fully epigenetic.
The environmental trigger to change the behaviour is
similar to an epigenetic mark, but is missing the ability to be
passed on and evolve in future generations. There are other algorithms
that aim to incorporate epigenetic mechanisms but do not meet the criteria of
high citation count or high performance in benchmarks. These algorithms
will be discussed next.

\section{Missing concepts from the Extended Evolutionary Synthesis}\label{sec:lack-of-epigenetics}
A few elements from the extended evolutionary synthesis have been
explored in successful algorithms: Swarm Intelligence algorithms
fit closely to the concepts of
cultural inheritance and information transfer; cMLSGA has been shown
to be a successful implementation of multilevel selection; Evolvability
is represented by hyper-parameter choices, which algorithms such as
CMA-ES dynamically alter during a search. Returning to the
biological concepts in Figure \ref{fig:synthesis-concepts}, this
leaves evo-devo theory, niche construction, epigenetic inheritance,
plasticity, and genomic evolution yet to be explored.
The improved performance of algorithms such as EBO with CMAR, Cuckoo-GRN,
GB-ABC, and co-evolutionary methods show that a merge of multiple
evolutionary techniques can have potential benefits.

While the extended synthesis provides a number of opportunities, not
all of the elements provide easy inspiration for practical optimisation
problems. Evolutionary and developmental biology (Evo-devo)
includes the developmental stages of living organisms and the evolution
of developmental processes. While in some ways most closely replicating
the initial vision proposed by Turing \cite{turing:1950} of generating
a child and teaching it to learn, adapting the concept
for general evolutionary algorithms would require substantial expansion of
the simple genotype model of evolution used in evolutionary
algorithms, such as including properties relating to gene regulation,
homeobox genes and allometry. Evo-devo provides a number of developmental
steps for fine tuning an organism, but in the algorithmic world this
is unnecessary when more generations can be run instead.
It is possible to apply evo-devo specific
applications, such as digital architectures \cite{evo-devo-algos},
where domain-specific concepts in architecture can be linked to evo-devo
processes, and the time taken to generate solutions means a lower
number of generations can be run. 
Similarly, genomic evolution involves the evolution of genome
architecture itself. In computational terms, genomic evolution would involve
the evolution of the number of variables or range of values in an optimisation
problem. These values are typically set based on the problem and do not
require evolutionary mechanics applied to them.

Niche construction requires individuals in the population to alter and
change their own environment to suit the population's properties. This allows
populations and genes that are normally less fit under the environment
to gain fitness due to the changes made to the environment, but
this is difficult to accomplish computationally. Optimisation problems
are usually predefined and the fitness landscape is dependent
on the problem. If an algorithm is able to change the fitness landscape
to suit the solutions it produces, it changes the problem definition,
meaning the solutions found may no longer be applicable or useful to
the original problem.

However, there are a range of unused mechanisms which can be considered to
provide excellent bio-inspiration for a new range of algorithms,
documented in the following subsections.

\subsection{Epigenetics}
Epigenetics plays an important role in adapting a population to new
conditions and environments quickly. In computational terms,
epigenetic mechanisms should help to improve convergence, potentially spreading
changes through a population faster than genetic evolution, and improve
stability around a solution in the face of environmental changes.
There are a number of different epigenetics processes such as DNA methylation,
bookmarking, gene silencing, gene repression, and genomic imprinting
that are triggered by different factors such as environment, diet, or the
presence of certain chemical compounds \cite{diabetes-epigenetics}.

In the Modern Synthesis view of evolution, genetic
changes occur randomly and fitness is guided by natural
selection. Similar to this, the development of modern genetic
algorithms focus on improving the selection process while
keeping genetic changes random to retain diversity. This
requires a number of generations for suitable traits with
high fitness to spread throughout a population, even with
mechanisms such as elitism. Epigenetic inheritance allows
for faster changes based on environmental cues which can occur
simultaneously among multiple individuals in the same generation.
These adaptive adjustments also do not affect the underlying
genotype, allowing regular genetic processes to occur and
epigenetic processes to be reversed.
The inheritance of epigenetic tags
in parallel with genetic inheritance results in continual
rapid changes with a diverse set of tags among a population,
without disrupting underlying genetic processes \cite{epigenetic-tags}.
This is an important aspect in evolutionary biology to
guide phenotypic variation in a direction suitable for the
environment instead of relying solely on random mutation
and natural selection.

\subsection{Phenotypic plasticity}
The concept of phenotypic plasticity is the idea that an organism's behaviour
or physiology could change due to environmental factors \cite{price-plasticity}.
The class of optimisation problems that focuses on dynamic
problems, where the objectives and constraints may be unstable and change
throughout the search, would be suitable for algorithms with flexible,
plastic responses. Algorithms in Swarm Intelligence can
exhibit some of this behaviour, for example when new bees take different roles
in the Artificial Bee Colony depending on the number of existing
bees in other roles. However, there is no scale or range of plastic responses
in reaction to a changing environment in the ABC implementation.

\section{Opportunities to include epigenetic mechanisms in evolutionary algorithms}
\label{sec:epigenetic-opportunities}

\subsection{Existing studies}
The key concept of epigenetics is to allow for fast variation when appropriate.
Existing studies \cite{periyasamy-ega, chrominski-ega, epiga}
inspired by epigenetics are currently missing the key feature
of triggering mechanisms based on the fitness of the population to
the environment. They often trigger the mechanism probabilistically without
any distinction between individuals, parents and how the epigenetic marks
are passed on. This probabilistic method is more akin to bet hedging \cite{bet-hedging}
than epigenetics, where the mechanics do not improve individual fitness in
stationary conditions with no drastic changes in the environment,
but create advantages in extreme conditions such as being stuck at a local optima.

An epigenetic algorithm
based on intra-generational epigenetic processes used by bio-molecules
is developed by Periyasamy \cite{periyasamy-ega}.
This system mimics cellular organisation, with individuals of bio-molecules
performing independent tasks in a swarm-like manner, and require specific
conditions to be met. 
The focus on the epigenetic processes uses no genetic
operators, which misses a genetic component to contribute to the
final phenotype. So while the use of epigenetics can improve the convergence
of this algorithm, it potentially loses diversity from the lack of genetics
and can get stuck at local optima.

\begin{figure}[H]
  \centering
  \includegraphics[width=0.7\textwidth, keepaspectratio]{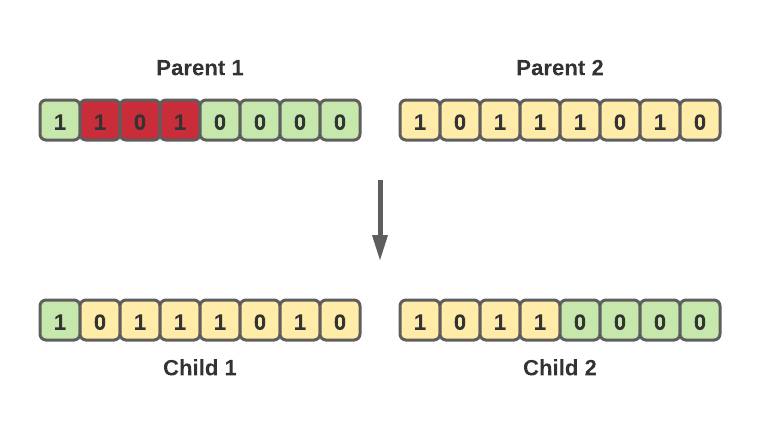}
  \caption{Concept of the cytosine methylation epigenetic process
    used in \cite{chrominski-ega}, where part of the parent genotype
    is blocked by the epigenetic process.}
  \label{fig:methylation-ga}
\end{figure}

The epigenetic process of
cytosine methylation is incorporated into a Genetic Algorithm
to solve the Knapsack problem \cite{chrominski-ega}.
Cytosine methylation blocks a
part of an individual's genotype during the crossover operation.
Figure \ref{fig:methylation-ga} shows the methylation operation used
in reproduction, where a portion of the crossover from parent 1 is blocked.
This aims to transfer a larger portion of the fitter parent's genotype
while silencing the poorer parent.
This constant probability for methylation to occur means the
epigenetic tags are assumed to always be passed to
new generations without any dynamic changes to external factors.

Finally, the concept of methylation
and gene silencing is incorporated into an epiGenetic Algorithm (epiGA)
\cite{epiga}.
Similar to \cite{chrominski-ega}, a number of parent genes may be masked based on
the probability of the epigenetics mechanism occurring. The constant
probability of trigger for the epigenetic mark means the marks do not evolve
or change through new generations.
The occurrence of the epigenetic trait is probabilistic,
so it occurs with some chance rather than
being triggered due to environmental or parental cues. This misses
the concept of allowing fast variation when appropriate.
So, while some algorithms are inspired by epigenetics they are currently
missing the key features that lead to the benefits seen in evolution.
These features are outlined in the following subsections.

%These existing studies on epigenetic mechanisms in genetic algorithms
%do not meet the criteria (\ref{apdx:algorithms-critera}) for
%either popularity or performance. Though they should improved performance
%against older algorithms, they have not been shown to compete against
%state-of-the-art algorithms in benchmarking competitions.

\subsection{Epigenetic tags - the epigenotype}
A key aspect existing studies do not capture is the inheritance
and transfer of epigenetic information to future generations.
While the epigenetic mechanisms implemented were accurate,
this transfer is important as it guides the direction
of phenotypic change. Without this aspect, the epigenetic
mechanisms simply act as another form of mutation, probabilistically
switching genes on and off.
To include the epigenetic information transfer,
epigenetic tags can be used to form an epigenotype \cite{epigenome},
to keep a history of inherited epigenetic changes and allow changes to
an individual's genotype to be triggered based on these tags.
Epigenetic tags can be added and removed based on signals
from the environment, or based on inheritance and crossover
operations when forming offspring individuals \cite{epigenetic-tools}.
Epigenetic tags can also help control gene
expression in response to environmental changes. In biology
this helps to form ``memory'' based on changing
environments \cite{epigenetic-memory}. The memory
of the recent environment allows for fast adaptation
and stability. By controlling how genes are expressed
using an epigenotype, suitable traits are constantly adjusted
to improve fitness before longer term genetic changes
can be applied.

\begin{figure}
  \centering
  \includegraphics[width=0.7\textwidth, keepaspectratio]{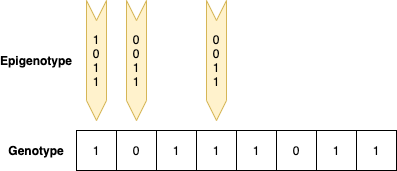}
  \caption{An epigenotype with epigenetic tags for some genes in
  the genotype.}
  \label{fig:epigenotype}
\end{figure}

Computationally, each variable in a solution can contain
a set of tags that can be inherited and modified.
Figure \ref{fig:epigenotype} shows the epigenetic
tags on top of some variables in the genotype.
The tag can be used to encode
mechanisms to alter the variable after genetic operators
are applied. These mechanisms can then help increase convergence
by guiding phenotypic variation in a direction matching
selection pressures. The underlying genetic mechanisms
are further unaffected by epigenetic changes, and
epigenetic changes can be easily reversed, reducing
the cost of poor mutations compared to genetic mutations.

The epigenotype represents the three key aspects of epigenetics:
\begin{itemize}
\item the transfer of epigenetic information,
\item self-adaptability to environmental changes,
\item fast convergence from direction variation.
\end{itemize}
These aspects may have potential benefits especially for dynamic problems
where the Pareto front is not static. The self-adaptability and fast convergence
would allow algorithms with an epigenotype to adjust to a changing
Pareto front quickly.

\subsection{Epigenetic mechanisms}
With the use of an epigenotype \cite{epigenome},
epigenetic tags can be added, removed,
and inherited to future generations. An epigenotype alters
the phenotype without changes to the underlying phenotype.
The tags can then be used
to encode different epigenetic mechanisms to be triggered.
The mechanisms have a range of effects on the genotype, such as
switching genes on and off, or reducing gene expression based
on the location and number of tags in the epigenotype.

\subsubsection{Genomic imprinting}
Genomic imprinting \cite{genomic-imprinting} restricts the expression
of a gene to one parent. Imprinting is useful when the imprinted alleles
lead to different phenotypes that affect an individual's fitness.
This process is non-Mendelian as it does not directly change the
genotype. Epigenetic tags are imprinted in the germline and cause
the imprinted genes to be expressed from only one parent
\cite{imprinting-germline}.

There are three hypothesised theories for the process of imprinting
\begin{itemize}
\item the kinship theory \cite{kinship-imprinting} -- The theory suggests
that an imbalance exists between parental genes due to conflicting
fitness strategies from both parents. This is mostly apparent in sexual
reproduction where the father and mother have differing interests
to pass on their own genes.
\item The sexual antagonism theory \cite{sexual-imprinting} -- This
  theory uses sex-specific selection pressure. It predicts an uneven
  allele frequency between males and females when natural selection
  favours one sex over the other so that offspring genes are
  enriched to benefit a particular sex.
\item The maternal–offspring coadaptation theory \cite{coadaptation-imprinting} --
  Based on the correlation between the genes of the mother and the maternal genes
  of the offspring, the maternal-offspring coadaptation theory states that the offspring
  is more likely to inherit from its mother because it leads to a higher
  probability that the offspring has a positive interaction with its maternal
  phenotype, and the interaction leads to higher fitness.
\end{itemize}

\subsubsection{Gene regulation}
The presence of epigenetic tags enables gene regulation mechanics
to occur on the tagged genes. There are multiple forms
of gene regulation: gene silencing, gene activation, and gene
repression. All forms of gene regulation affect the
expression of the affected genes leading to variation in
phenotypes from the same genotype \cite{gene-regulation-mammals}.
Gene silencing is a mechanism for turning entire sections of
the genotype on and off independent of mutation.
In evolutionary biology
gene silencing has the effect of protecting the host
organism from viruses \cite{rna-silencing} by silencing
genes that are used in viral reproduction. In terms of
convergence and diversity, convergence should be increased
and diversity decreased as silenced genes are not fully
expressed compared to other genes.

Gene repression acts on individual genes rather than entire
sections of the genotype. In evolutionary biology it
switches off genes whose products are required to maintain
cell functions \cite{cellular-aging}.
To implement this computationally, each variable in a
candidate solution can be switched on or off,
based on the tags of the epigenotype. The modification
of the tags can be based on the fitness of the individual, adapted based on
the progress of the search or based on environmental cues.

\section{Conclusion}
Evolutionary Computation is an area that now covers many hundreds of different
variants. Despite its biological roots, and the diversity of mechanisms
available in nature, these algorithms can broadly be split
into two main categories of evolution: Evolutionary Algorithms, which take
inspiration from genetic inheritance, and Swarm Intelligence algorithms,
which take inspiration from cultural inheritance. The reason for this lack of
diversity is speculated to be that similar underlying biological mechanisms are
often recycled under new banners, making it hard to understand which mechanisms
have already been considered. To understand which have been considered, and
which have been overlooked, existing Evolutionary Computation algorithms are
categorised under a contemporary biological classification inspired by the
Extended Evolutionary Synthesis. It confirms that two main types
of information exchange, and therefore behaviours, are used in Evolutionary Computation.
Few algorithms even go as far as combining
these types of information exchange but with those that do providing excellent performance.

The framework shows that the existing concepts
from the Modern Evolutionary Synthesis have already been incorporated and that
some concepts of the Extended Evolutionary Synthesis have also recently been
included, with benefits to performance of those algorithms.
Opportunities for Evolutionary Computation with
epigenetic mechanisms are identified, which is a key element of the way
evolution is now described with many underlying mechanisms proposed in the
biological literature. These mechanisms provide a self-adaptive means to quickly
converge and stabilise populations in changing environments.
A key aspect in inspiring algorithms from epigenetics is flagged as the transfer
of epigenetic information to quickly change individual phenotypes,
which must be reversible so that they do not alter the genotype.
The resulting mechanisms will be both self-adaptive
and convergence based, which could provide benefits to Evolutionary
Computation.

\section*{Acknowledgement}
The authors would like to thank the Lloyds Register Foundation and the Southampton Marine
and Maritime Institute for their kind support of this research.

\bibliography{references}

\newpage
\appendix
\section{Algorithms chosen}\label{apdx:algorithms-criteria}
\begin{table}[H]
  \caption{Evolutionary algorithms were chosen if they had more than 1000 citations
    on Google Scholar, or showed exceptional performance in a competition or benchmarking
  paper.}
\begin{center}
  \begin{tabular}{| m{2cm} | m{1.5cm} | m{8cm} |}
    \hline
    \textbf{Algorithm} & \textbf{Citations} & \textbf{Reason} \\ \hline

    \rowcolor{ec}
    NSGA-II & 35038 &
    Most popular multi-objective genetic algorithm with over 1000 citations
    and used to represent the niching based family of multi-objective
    genetic algorithms. \\ \hline

    \rowcolor{ec}
    MOEA/D & 5205 & Original algorithm in the decomposition based family of multi-objective
    genetic algorithms. Over 1000 citations. \\ \hline

    \rowcolor{ec}
    IBEA & 1925 & An indicator based genetic algorithm using an indicator for selection,
    which contrasts with the common Pareto dominance ranking used for selection.
    Over 1000 citations.\\ \hline

    \rowcolor{ec}
    HEIA & 94 & One of the top performing multi-objective algorithm from
    recent \cite{slaw-cmlsga} benchmarking on 100 Evolutionary Computation
    problems representing the co-evolutionary family of genetic algorithms. \\ \hline

    \rowcolor{ec}
    cMLSGA & 6 & One of the top performing multi-objective algorithm from
    recent \cite{slaw-cmlsga} benchmarking on 100 Evolutionary Computation
    problems representing the multi-level selection family of genetic algorithms. \\ \hline

    \rowcolor{ec}
    DE & 25723 & Original differential evolution algorithm which many CEC single objective
    competition winners are based off of. Over 1000 citations. \\ \hline

    \rowcolor{ec}
    SHADE & 614
    & An extension of DE with improved performance using a memorised history
    of successful parameters. SHADE variants have won or come
    top in multiple CEC competitions from 2014 to 2019. \\ \hline

    \rowcolor{ec}
    CMA-ES & 1858 & Algorithm from the family of Evolution Strategies
    algorithms with enhanced performance over the original ES algorithm.
    Over 1000 citations. \\ \hline

    \rowcolor{si}
    CLPSO & 3253 & Popular variant of the particle swarm optimisation algorithm.
    Over 1000 citations. \\ \hline

    \rowcolor{si}
    ACO & 13431 & Algorithm used primarily for path
    finding and routing problems. Over 1000 citations.\\ \hline

    \rowcolor{si}
    ABC & 6625 & Algorithm based on the foraging behaviour of bees.
    Over 1000 citations.\\ \hline

    \rowcolor{si}
    EBO with CMAR & 67 &
    Top performance in CEC'17 Bound Constrained Competition \cite{cec-17}. \\ \hline

    \rowcolor{si}
    FA & 3533 & Based on the flashing behaviour of fireflies.
    Over 1000 citations. \\ \hline

    \rowcolor{si}
    Cuckoo Search & 5553 & Based on the Cuckoo species laying eggs
    in the nests of other birds. Over 1000 citations. \\ \hline

    \rowcolor{si}
    GWO & 5137 & Algorithm inspired by the hunting behaviour of grey wolves.
    Over 1000 citations. \\ \hline
  \end{tabular}
\end{center}
\end{table}

\end{document}